%% file: main.tex
\documentclass[letterpaper, 10pt, conference]{ieeeconf}  
\usepackage{xcolor}

\IEEEoverridecommandlockouts                              

\usepackage{xr}

\newcommand{\omitted}[1]{}

\input{titleAndAuthors}

\usepackage{cite}

\usepackage{comment}
\usepackage{siunitx}
\usepackage{relsize}
\usepackage{ifthen}
\usepackage[colorinlistoftodos]{todonotes}
\usepackage[vlined,ruled,linesnumbered]{algorithm2e}
\usepackage{graphics} 
\usepackage{rotating}
\usepackage{color}
\usepackage{enumerate}
\usepackage[T1]{fontenc}
\usepackage{psfrag}
\usepackage{epsfig} 
\usepackage{booktabs}
\usepackage{graphicx}
\usepackage{url}
\usepackage{multirow}
\usepackage{array}
\usepackage{latexsym}
\usepackage{amsfonts}
\usepackage{amsmath}
\DeclareMathOperator{\tr}{tr}
\usepackage{gensymb}
\usepackage{amssymb}
\usepackage{xstring}
\usepackage{algorithmic}
\usepackage{multirow}
\usepackage{xcolor}
\usepackage{prettyref}
\usepackage{flexisym}
\usepackage{bigdelim}
\usepackage{breqn} 
\usepackage{listings}
\let\labelindent\relax
\usepackage{xspace}
\usepackage{float}
\usepackage{bm}
\usepackage{soul}
\graphicspath{{./figures/}}
\usepackage{tikz}
\usetikzlibrary{matrix,calc}
\usepackage{lipsum}
\usepackage{mdwlist}

\let\itemize\undefined
\makecompactlist{itemize}{stditemize}
\usepackage{enumitem}
\usepackage{caption}
\usepackage{epstopdf}
\usepackage{subfigure}
\usepackage{svg}

\usepackage{mathtools}

\makeatletter
\let\NAT@parse\undefined
\makeatother
\usepackage{hyperref}
\usepackage{cleveref}

\hypersetup{%
  colorlinks=true,
  linkcolor=blue,
  filecolor=magenta,      
  urlcolor=black,
  citecolor=red,
  linkbordercolor={0 0 1}
}

\input{preamble_symbols.tex}

\input{shortcuts}

\PassOptionsToPackage{end}{algorithmic}

\renewcommand{\baselinestretch}{.98}

\input{titleAndAuthors}
\begin{document}

\maketitle
\thispagestyle{empty}
\pagestyle{empty}

\input{0-abstract}
\input{1-intro}

\input{2-dynamics}
\input{3-control}

\input{4-inverse}

\input{5-exp}

\input{6-conclusion}

\bibliographystyle{IEEEtran}
\bibliography{references}

\end{document}

%% file: titleAndAuthors.tex
\title{\LARGE
\textbf{MorphEUS: Morph}able Omnidir\textbf{E}ctional \textbf{U}nmanned \textbf{S}ystem\textsuperscript{\textdagger}}

\author{Ivan Bao$^\star$, José C.~Díaz Peón González Pacheco$^\star$, Atharva Navsalkar$^\star$,\\ Andrew Scheffer$^\star$, Sashreek Shankar$^\star$,
Andrew Zhao$^\star$, Hongyu Zhou$^\star$, and Vasileios Tzoumas
\thanks{\textsuperscript{\textdagger}This work was presented at the Workshop on Open Challenges in Robotics for Asset Inspection and Management (OCRAIM) at the 2025 IEEE International Conference on Robotics and Automation (ICRA).}%
\thanks{$^\star$Authors ordered alphabetically. A.Z. and I.B. contributed to the hardware design. A.S., S.S., and J.D. contributed to the firmware and simulations. H.Z. contributed to the geometric control. A.N. contributed to the control allocation.  A.S., S.S., J.D., A.N., and H.Z. contributed to the writing.}
\thanks{This work was conducted at the University of Michigan with the partial support of Undergraduate Research Opportunity Program (see the above link to featured media) and the Michigan Translational Research and Commercialization (MTRAC) for Advanced Transportation Innovation Hub through Kickstart Early-Stage Funding program.}
}

%% file: preamble_symbols.tex
\newtheorem{corollary}{Corollary}

\newtheorem{proposition}{Proposition}

\newtheorem{remark}{Remark}

\newcommand{\bdmath}{\begin{dmath}}
\newcommand{\edmath}{\end{dmath}}
\newcommand{\beq}{\begin{equation}}
\newcommand{\eeq}{\end{equation}}
\newcommand{\bdm}{\begin{displaymath}}
\newcommand{\edm}{\end{displaymath}}
\newcommand{\bea}{\begin{eqnarray}}
\newcommand{\eea}{\end{eqnarray}}
\newcommand{\beal}{\beq \begin{array}{lll}}
\newcommand{\eeal}{\end{array} \eeq}
\newcommand{\beas}{\begin{eqnarray*}}
\newcommand{\eeas}{\end{eqnarray*}}
\newcommand{\ba}{\begin{array}}
\newcommand{\ea}{\end{array}}
\newcommand{\bit}{\begin{itemize}}
\newcommand{\eit}{\end{itemize}}
\newcommand{\ben}{\begin{enumerate}}
\newcommand{\een}{\end{enumerate}}

\newcommand{\calB}{{\cal B}}

\newcommand{\calM}{{\cal M}}
\newcommand{\calN}{{\cal N}}

\newcommand{\calP}{{\cal P}}

\definecolor{myblue}{RGB}{65 105 225}

\newcommand{\hide}[1]{}

\newcommand{\hiddenText}{{\color{gray} hidden text.}}
\newcommand{\hideWithText}[1]{\hiddenText}



%% file: shortcuts.tex
\newcommand{\ie}{\emph{i.e.},\xspace}


%% file: 0-abstract.tex
\begin{abstract}
Omnidirectional aerial vehicles (OMAVs) have opened up a wide range of possibilities for inspection, navigation, and manipulation applications using drones. In this paper, we introduce MorphEUS, a morphable co-axial quadrotor that can control position and orientation independently with high efficiency. It uses a paired servo motor mechanism for each rotor arm, capable of pointing the vectored-thrust in any arbitrary direction. As compared to the \textit{state-of-the-art} OMAVs, we achieve higher and more uniform force/torque reachability with a smaller footprint and minimum thrust cancellations.
The overactuated nature of the system also results in resiliency to rotor or servo-motor failures. The capabilities of this quadrotor are particularly well-suited for contact-based infrastructure inspection and close-proximity imaging of complex geometries. In the accompanying control pipeline, we present theoretical results for full controllability, almost-everywhere exponential stability, and thrust-energy optimality. We evaluate our design and controller on high-fidelity simulations showcasing the trajectory-tracking capabilities of the vehicle during various tasks. 
Supplementary details and experimental videos are available on the project webpage.\\
\\
Website: \url{https://iral-morphable.github.io}\\
Featured media (2022): \url{https://youtu.be/wFEDdiamtT8}

\end{abstract}

%% file: 1-intro.tex
\newcommand{\fixme}[1]{\textcolor{blue}{#1}}

\section{Introduction}\label{sec:intro}

\begin{figure}[t!]
    \centering
    \includegraphics[width=0.45\textwidth]{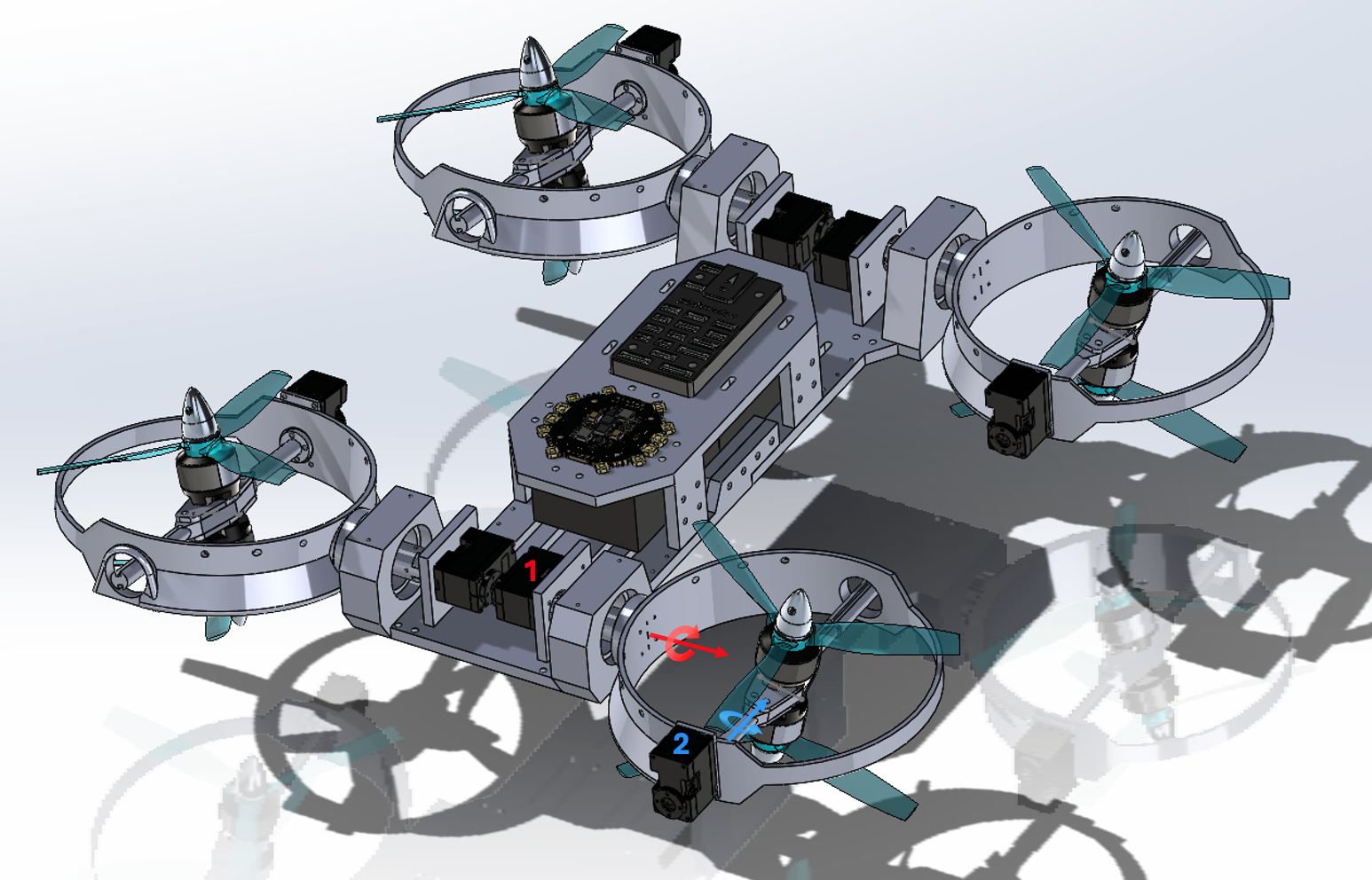}
    \caption{CAD of the proposed morphable quadrotor actuated with two additional degrees of freedom for each rotor, providing $360^{\circ}$ thrust-vectoring capabilities.}
    \label{fig:morphology}
    \vspace{-4mm} 
\end{figure}

Multirotor aerial vehicles have seen tremendous applications in the real world, including industrial inspection~\cite{seneviratne2018smart}, search and rescue~\cite{waharte2010supporting}, and automated crop monitoring~\cite{cuaran2021crop}. 
Standard multirotors suffer from limited maneuverability due to their limited controllability: the control of their position and orientation is coupled. This under-actuation heavily compromises their deployment in contact-based inspections, cluttered environments, and sensing and mapping applications. Omnidirectional multirotors (OMAVs) are aerial robots that enhance maneuverability through over-actuation and, in some cases, adaptive rotor configurations. This work introduces a novel OMAV (see \Cref{fig:morphology} for CAD design; \Cref{fig:real_world} for hardware prototype) capable of pointing the thrust in arbitrary directions. As compared to the \textit{state-of-the-art}, we achieve higher and more uniform force/torque reachability with a smaller footprint and minimum thrust cancellations.

\subsection{Related Work}
As multirotor robotic platforms have become more ubiquitous, there has been an increasing demand to extend their maneuverability \cite{active_interaction, past_present_future}. Many works in the last decade have focused on developing fully-actuated multirotors that have at least some decoupling of position and orientation. The current literature of fully-actuated multirotors can be divided into two broad classes: 1) \textit{variable-tilt multirotors}, which can actuate the orientation of their rotors during flight, and 2) \textit{fixed-tilt multirotors}, which cannot \cite{rashad2020fully}.

\textit{Fixed-tilt multirotors} like \cite{fixed_2jiang2013hexrotor,fixed_3rajappa2015modeling,fixed_4lei2017aerodynamic} have rotors mounted at fixed angles relative to the chassis.
While this simplifies the mechanical design, it often limits omnidirectional motion, restricts versatility to specific tasks such as constrained navigation, and reduces efficiency due to thrust cancellations. The octorotor configuration in \cite{brescianini2016design} achieves omnidirectional motion using out-of-plane rotor placements enclosed inside a cube. However, it suffers from inefficient force cancellations, no payload capacity, and a large footprint.  

\textit{Variable-tilt multirotors}, such as \cite{ryll2015novel, segui2014novel}, enable synchronized thrust vectoring for quadrotors, allowing more versatile motion control. However, they face limitations in achieving full force-torque reachability and often impose strict orientation-tracking constraints. The Voliro hexacopter \cite{kamel2018voliro, bodie2020towards}, utilizing six independent servos, is one of the most advanced fully-actuated UAVs, offering omnidirectional control. Despite its capabilities, it still faces challenges related to complex mechanics and limited efficiency in certain maneuvers. Similarly, \cite{aboudorra2024modelling} proposes a fully omnidirectional morphable octorotor with rotors that tilt synchronously. This design offers enhanced flexibility but still struggles with thrust inefficiencies and mechanical complexity.

\begin{figure}[t]
    \centering
    \includegraphics[width=0.49\textwidth]{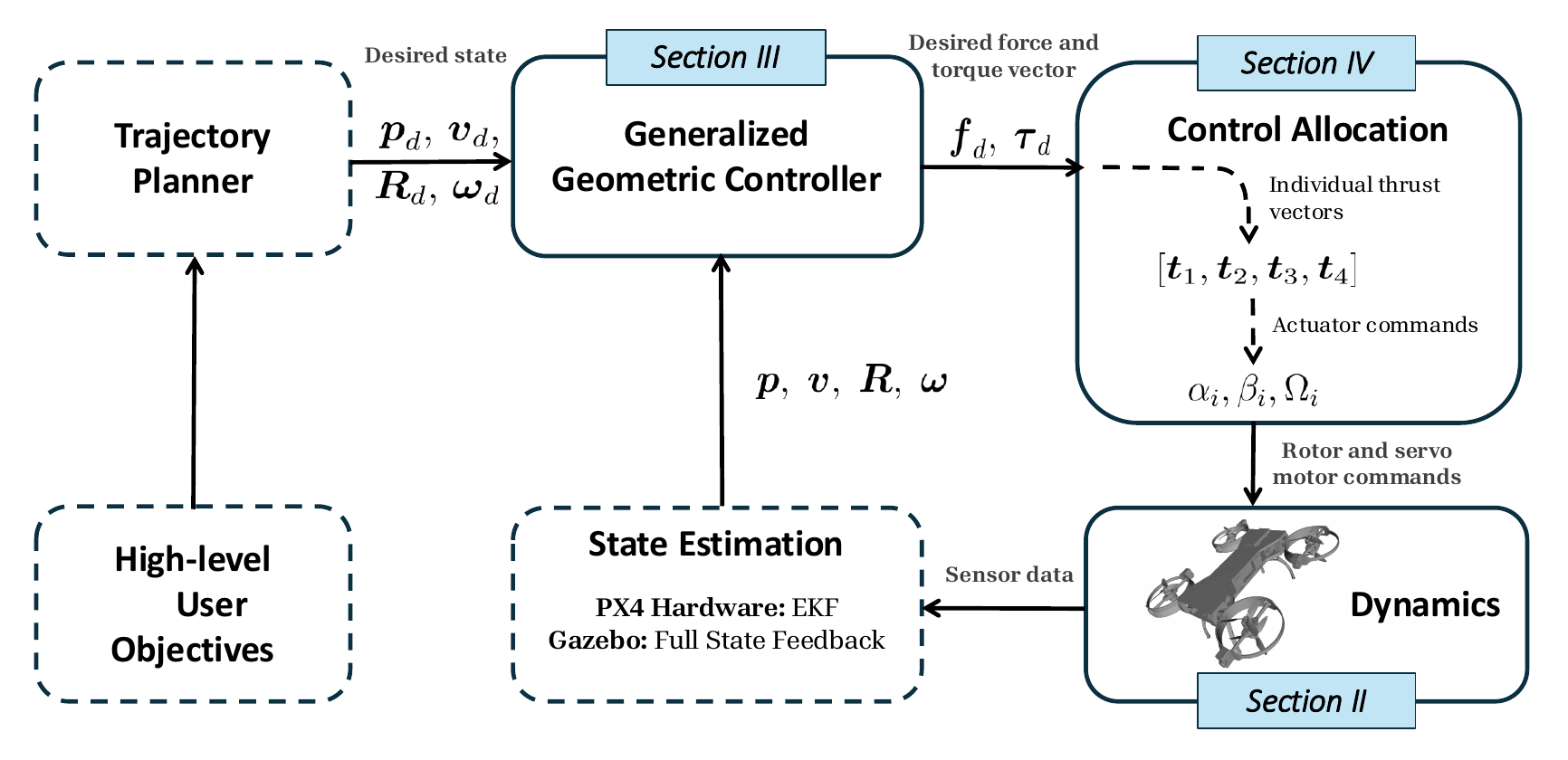}
    \caption{Autonomous pipeline for the proposed variable-tilt morphable quadrotor.}
    \label{fig:pipeline}
    \vspace{-4mm} 
\end{figure}

\subsection{Contributions}

We present a novel variable-tilt co-axial quadrotor (\Cref{fig:morphology}) and a control pipeline (\Cref{fig:pipeline}) that achieves a significantly higher level of maneuverability, dexterity, and efficiency. This system builds on prior work by improving omnidirectional flight capabilities while avoiding inefficient thrust cancellations. To realize this, each of the rotor pairs can be independently oriented using a pair of servo motors (\Cref{fig:morphology}). This allows for full 6-DoF control of position and orientation. We prove that the proposed controller is exponentially stable \textit{almost} everywhere and energy-optimal in terms of control allocation. In comparison to existing literature, our contributions are on three broader aspects:

\subsubsection{Dexterity and Full Omnidirectionality}
Our quadrotor can achieve all the force-torque realizations (within actuator limits) in all directions. We prove the system is fully controllable and reachable, enabling our drone to traverse complex and constrained geometries. This capability enhances dexterity for inspection and manipulation tasks.

\subsubsection{Aggressive Maneuverability}
Unlike \cite{aboudorra2024modelling}, our design achieves omnidirectionality with a much smaller footprint, resulting in higher maneuverability. The proposed geometric controller, generalizing \cite{lee2010geometric}, has an \textit{almost} $360^{\circ}$ region of attraction, enabling the vehicle to achieve exponential stability at almost any configuration in $SO(3)$. 

\subsubsection{High Efficiency with Resiliency}
Despite being an overactuated system, our control allocation scheme has zero thrust cancellation for translation, compared to \cite{bodie2020towards}. For general force-torque commands, we propose an intuitive and theoretically energy-optimal control allocation strategy. Moreover, our quadrotor design employs co-axial rotors, negating the rotor counter-torques and increasing thrust density. This design is robust for up to three rotor failures due to independent thrust vectoring at each rotor arm. 

\Cref{fig:pipeline} shows the proposed control pipeline for our quadrotor. We first discuss the dynamics of the quadrotor based on the morphing rotors (actuators) in \Cref{sec:dynamics}. There, we expand on the controllability and reachability propositions. Next, in \Cref{sec:control}, we present the generalized geometric controller and the results on exponential stability. The low-level control allocation scheme is elaborated in \Cref{sec:inverse}. Finally, we validate our proposed system using three high-fidelity simulation experiments that demonstrate the 6-DoF tracking capabilities of the quadrotor in \Cref{sec:exp}.

%% file: 2-dynamics.tex
\section{Morphable Quadrotor Dynamics}\label{sec:dynamics}

\begin{figure}[t]
    \centering
    \includegraphics[width=0.8\columnwidth]{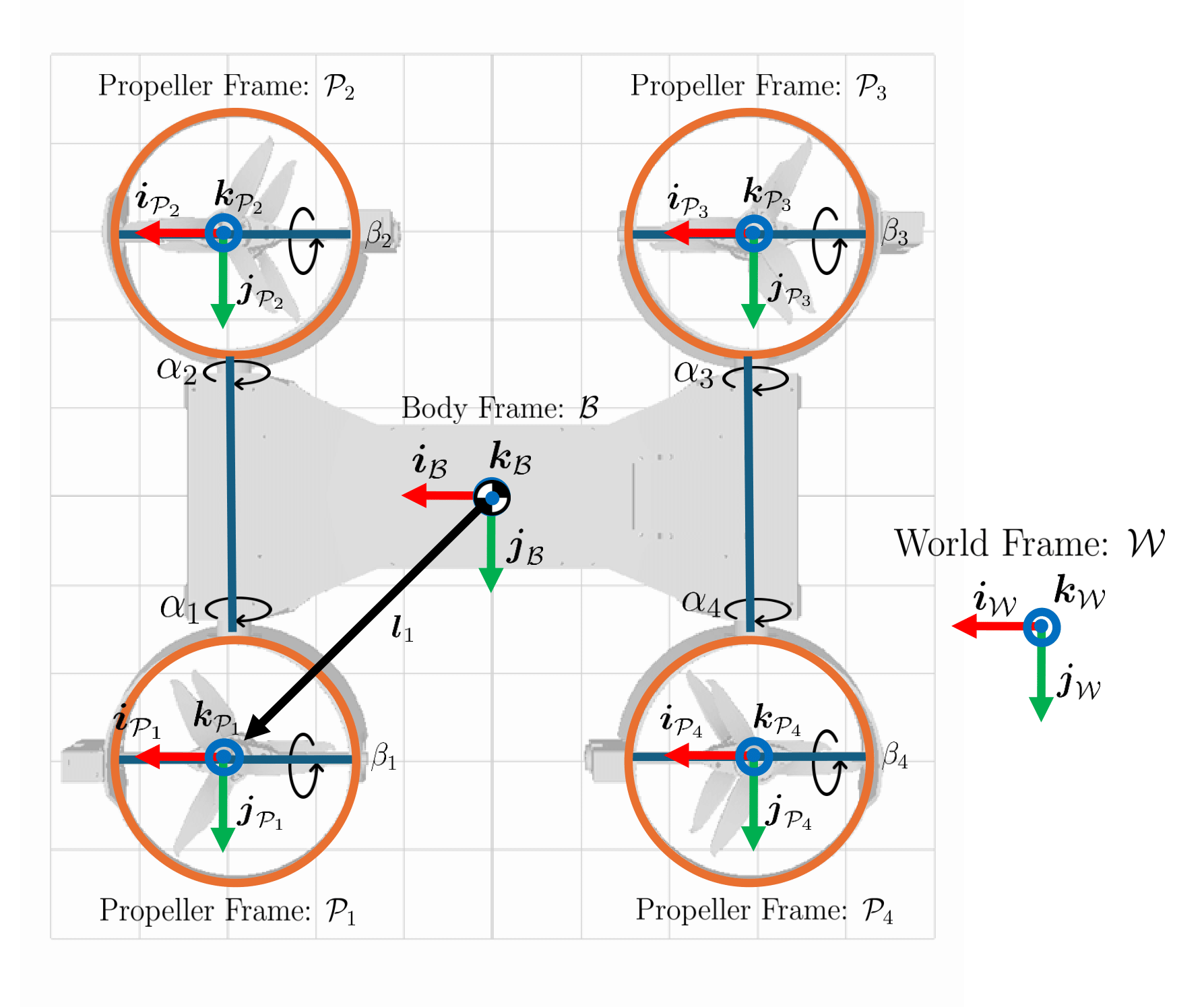}
    \caption{Schematic of the drone highlighting coordinate frames and servo angles.}
    \label{fig:frames}
    \vspace{-4mm} 
\end{figure}

In this section, we introduce the morphable quadrotor dynamical model. The model consists of the translational dynamics and the rotational dynamics. We use ${\mathcal{W}}$ to denote the world frame, ${\mathcal{B}}$ to denote the body-fixed frame, and ${\mathcal{P}}$ to denote the propeller-fixed frame, shown in \Cref{fig:frames}. We use a $Z-X-Y$ Euler angle sequence for yaw $\psi$, roll $\phi$, and pitch $\theta$ to represent the orientation. By design, the \textit{body} to \textit{propeller} frame transformation is characterized by two angles $\alpha$ and $\beta$, as shown in \Cref{fig:frames}. Each of these angles is controlled by a servo motor, providing additional actuator commands alongside the propeller speeds.

\subsection{Translational Dynamics}

The translational dynamics of the quadrotor is given by
\begin{equation}
    \begin{aligned}
        \dot{\boldsymbol{p}} = \boldsymbol{v}, \quad m \dot{\boldsymbol{v}} = m \boldsymbol{g} + \boldsymbol{R} \boldsymbol{f},
    \end{aligned}    
    \label{eq:translational_dynamics}
\end{equation}
where $\boldsymbol{p} \in \mathbb{R}^{3}$ and $\boldsymbol{v} \in \mathbb{R}^{3}$ are position and velocity in the world frame,
$m$ is the quadrotor mass, $\boldsymbol{g}\in \mathbb{R}^{3}$ is the acceleration due to gravity in the world frame, $\boldsymbol{R}\in SO(3)$ is the rotation matrix from body to world frame, $\boldsymbol{f}\in \mathbb{R}^{3}$ is the body forces generated by propellers in the body frame. 

Our design enables the generation of body forces $\boldsymbol{f}$ in any direction and it can be expressed as
\begin{equation}
\begin{aligned}
    \boldsymbol{f} = \sum_{i=1}^{4} (c_t \Omega_i^2)\boldsymbol{R}_{\mathcal{P}_i/\mathcal{B}}(\alpha_i,\beta_i)  \boldsymbol{k}_{\mathcal{P}_i} = c_{t} \boldsymbol{G} \boldsymbol{\Omega}^{\degree2},
    \label{eq:force_omega_relation}
\end{aligned}
\end{equation}
where $\boldsymbol{R}_{\mathcal{P}_i/\mathcal{B}}(\alpha_i,\beta_i)$ is the body-to-propeller rotation matrix,  $\boldsymbol{\Omega} = \left[ \Omega_1, \; \Omega_2, \; \Omega_3, \; \Omega_4\right]^\top$ is the angular rates of propellers, $c_{t}$ is the thrust coefficient, $\degree$ indicates the Hadamard power, and $\boldsymbol{G}$ is a suitable matrix that maps angular rates of propellers to body forces.
    
\vspace{\baselineskip}

\subsection{Rotational Dynamics}
The rotational dynamics of the quadrotor is given by
\begin{align}
    \dot{\boldsymbol{R}} = \boldsymbol{R}[{\boldsymbol{\omega}}]_\times, \quad & \mathcal{J} \dot{\boldsymbol{\omega}} = -\boldsymbol{\omega} \times \mathcal{J} \boldsymbol{\omega} + \boldsymbol{\tau}, \label{eq:rotational_dynamics}
\end{align}
where $\boldsymbol{\omega}\in\mathbb{R}^3$ is the angular velocity in the body frame, $[\cdot]_\times: \mathbb{R}^3\rightarrow\mathbb{R}^{3\times 3}$ is the skew-symmetric matrix operator,
$\mathcal{J}\in\mathbb{R}^{3\times3}$ is the inertia matrix in the body frame, and $\boldsymbol{\tau}\in\mathbb{R}^3$ is the torque generated by propellers in the body frame. 

The first term in the angular velocity dynamics is the gyroscopic moment, due to the change in the angular momentum. In the co-axial configuration, we constrain the rotational speed of both the propellers on one arm to be equal. This negates the drag force (angular) of each of the propellers due to equal and opposite spinning co-axial propellers. Since the net angular momentum of the individual pair is zero, it is energy efficient to point the thrust in any direction, using minimal effort from servo motors. The torque $\boldsymbol{\tau}$ is only a function of the individual thrust vectors, given by
\begin{align}
    \boldsymbol{\tau} = \sum_{i=1}^{4} (c_t \Omega_i^2) \boldsymbol{l}_i\times\hat{\boldsymbol{t}}_i, \label{eq:torque_thrust_relation}
\end{align}
where \(\boldsymbol{l}_i = l_{ix} \boldsymbol{i}_\calB + l_{iy} \boldsymbol{j}_\calB + l_{iz} \boldsymbol{k}_\calB\) is the position vector of the center of the \(i\)-th propeller with respect to center of mass (COM) of the quadrotor in the body frame, \(\hat{\boldsymbol{t}}_i\in\mathbb{R}^3\) is the individual thrust direction (unit vector) for each propeller in the body frame. Mathematically, it is expressed by \(\hat{\boldsymbol{t}}_i=\boldsymbol{R}_{\mathcal{P}_i/\mathcal{B}} (\alpha_i, \beta_i)\boldsymbol{k}_{\calP_i}\). Substituting these relations, the torque exerted by the propellers is given by
\begin{equation}
\begin{aligned}
    \boldsymbol{\tau} &= \sum_{i=1}^{4} (c_t \Omega_i^2) \boldsymbol{l}_i\times\boldsymbol{R}_{\mathcal{P}_i/\mathcal{B}} (\alpha_i, \beta_i)\boldsymbol{k}_{\calP_i} = c_{t} \boldsymbol{H} \boldsymbol{\Omega}^{\degree2},
    \label{eq:torque_omega_relation}
\end{aligned}
\end{equation}
where $\boldsymbol{H}$ is a suitable matrix that maps angular rates of propellers to body moments.

\vspace{\baselineskip}
Now we have a relation that gives both force and torque applied on the quadrotor due to the angular velocities and direction (based on \(\alpha_i, \beta_i\)) of the propellers. Combining the translation and rotational force equations, we have 
\begin{align}
    \begin{bmatrix}
        \boldsymbol{f}\\ \boldsymbol{\tau}
    \end{bmatrix} = c_t\begin{bmatrix}
        \boldsymbol{G} \\ \boldsymbol{H}
    \end{bmatrix}\boldsymbol{\Omega}^{\degree2} = c_t \boldsymbol{F}(\alpha_i,\beta_i)\boldsymbol{\Omega}^{\degree2}.\label{eq:F_matrix}
\end{align}

This is a forward dynamics relation that gives the force and torque exerted due to control inputs $\alpha_i$, $\beta_i$ and $\Omega_i$. 

In \Cref{sec:inverse}, we present the inverse relation, that is, the control allocation to compute the values of $\alpha_i$, $\beta_i$ and $\Omega_i$, for the desired reference force \(\boldsymbol{f}_d\) and torque \(\boldsymbol{\tau}_d\). The novel configuration of our quadrotor allows us to command omnidirectional force and torque independently, subject only to the saturation limit of the individual propeller pair.

\begin{figure}[t]
    \centering
    \includegraphics[width=1\linewidth]{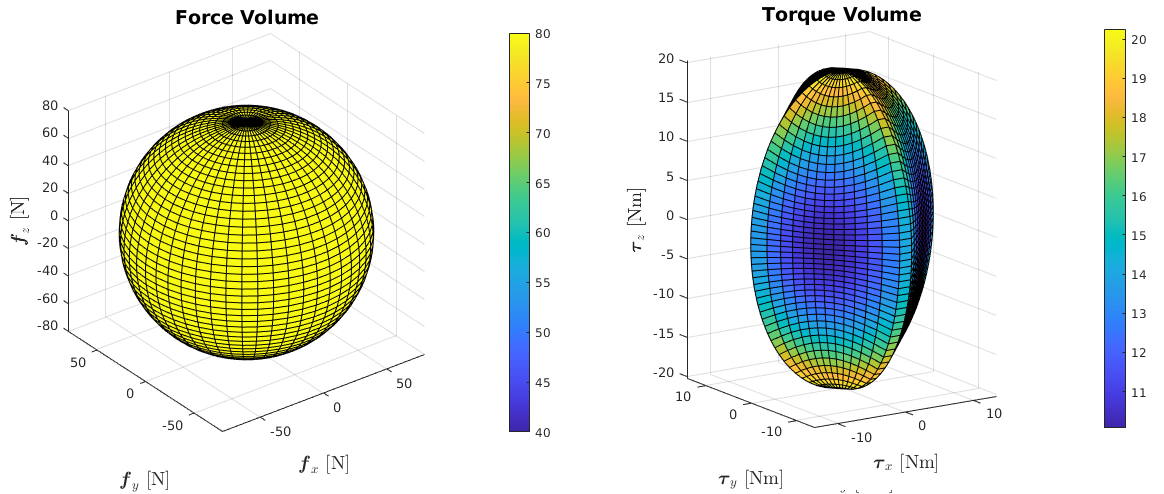}
    \caption{Force and torque envelopes, assuming that each arm can apply a maximum thrust of 20N. Force envelope assumes desired torque to be zero, and vice versa.}
    \label{fig:envelope}
    \vspace{-4mm} 
\end{figure}

\subsection{Reachability and Controllability}
We discuss that our vehicle is fully controllable and can command omnidirectional force and torque independently.
We omit proofs, while highlighting the broad methodology. 
\begin{proposition}[Reachability]
    Given the drone dynamics \eqref{eq:translational_dynamics} and \eqref{eq:rotational_dynamics}, force and torque relations \eqref{eq:F_matrix} and \textit{sufficient} limits on rotor velocities,  all the states $\mathbf{x} = [x,y,z,\dot{x},\dot{y},\dot{z},\phi,\theta,\psi,p,q,r,]^\top$ (position, linear velocity, orientation, angular velocity) are reachable.
\end{proposition}
To prove this, we first consider the concatenation of individual thrust vectors \(\boldsymbol{t}=\begin{bmatrix}\boldsymbol{t}_1&\boldsymbol{t}_2&\boldsymbol{t}_3&\boldsymbol{t}_4 \end{bmatrix}^\top\in \mathbb{R}^{12}\) as the virtual control input. Intuitively, we can always find values of actual control inputs $\alpha_i,\beta_i$ and $\Omega_i$ that result in any individual thrust vector $\boldsymbol{t}_i$, given that rotor speed $\Omega_i$ goes sufficiently high. 
We can then find that the force-torque vector $[\boldsymbol{f},\boldsymbol{\tau}]^\top \in \mathbb{R}^6$ can be mapped from the virtual control input $\boldsymbol{t}\in \mathbb{R}^{12}$ through a linear mapping function $\calM:\mathbb{R}^{12}\rightarrow \mathbb{R}^6$.
We then show that this has infinite solutions, implying that all forces and torques are possible, and hence all the states are reachable. 

\begin{corollary}[Controllability]
    Since the solution for the linear map $\mathcal{M}$ always exists, we can always find a set of thrust vectors $\vec{t}_i$, such that a desired force and torque is applied. Mathematically, given vectors $\boldsymbol{f}$ and $\boldsymbol{\tau}$ there exists $\boldsymbol{t} = \begin{bmatrix}\boldsymbol{t}_1&\boldsymbol{t}_2&\boldsymbol{t}_3&\boldsymbol{t}_4 \end{bmatrix}^\top$ such that $\boldsymbol{f}=\sum_{i=1}^4 \boldsymbol{t}_i, \boldsymbol{\tau}=\sum_{i=1}^4 \boldsymbol{l}_i\times\boldsymbol{t}_i$. Hence, the system is \textit{controllable} as all force and torque realizations are possible. 
\end{corollary}


Accounting for rotor speed limits, we get a range of feasible force and torque values under the individual thrust saturation assumption.  \Cref{fig:envelope} shows the possible force and torque vector envelope.
The force envelope assumes that the commanded torque is zero and vice versa. We see that the force limits are uniform in all directions, while the torque limits have a minimum-to-maximum limit ratio of roughly \(0.5\). Particularly, the uniform force reachability  across all directions holds a significant advantage over existing state-of-the-art designs such as the VoliroX hexacopter \cite{bodie2020towards,allenspach2020design,bodie2023omnidirectional}, where the force envelope is non-uniform.

\begin{remark}[Gimbal Lock]
    We emphasize that our quadrotor design has capability to exert force and torque, that is, wrench in any given direction. While this is true in any static state, the controllability is constrained in certain directions, when the angle $\beta=\pm\pi/2$ due to the presence of gimbal lock. This aspect is one of the inherent limitations of the design choice where each arm has a pair of orthogonally placed servo motors (angles $\alpha_i$ and $\beta_i$). The quadrotor has the control authority to command the angular velocity $\Omega_i$ along $\boldsymbol{k}_{\calP_i}$ axis, and angles $\alpha_i,\beta_i$ along  $\boldsymbol{j}_{\calB}$ and $\boldsymbol{i}_{\calP_i}$ axes respectively, as seen in \Cref{fig:frames}. When $\beta_i=\pm\pi/2$ i.e., desired force $\boldsymbol{f}_d$ is along $\boldsymbol{j}_{\calB}$, we see that axes $\boldsymbol{k}_{\calP_i}$ and $\boldsymbol{j}_{\calB}$ align with each other. This is why we lose the control authority along yaw axis $\boldsymbol{k}_{\calB}$ as we cannot command any thrust in $\boldsymbol{i}_{\calB}-\boldsymbol{j}_{\calB}$ plane (only in $\boldsymbol{j}_{\calB}$ direction). We can resolve this issue by introducing a differential thrust controller for yaw angle. During the configuration of the gimbal lock, we can appropriately create a differential in thrust vector magnitudes along $\boldsymbol{j}_{\calB}$ axis to exert a moment along $\boldsymbol{k}_{\calB}$ axis. In practice, we will implement this as a hybrid control strategy when $\beta_i\approx\pm\pi/2$.
\end{remark}

%% file: 3-control.tex
\section{Geometric Control}\label{sec:control}
We present a generalized geometric controller for the morphable quadrotor to track desired positions and orientations.
The controller aims to calculate the desired force and torque commands in 6-DOF, based on the tracking error in position and orientation.
In contrast to the geometric controller of a regular quadrotor~\cite{lee2010geometric}, the generalized geometric controller of the morphable quadrotor obtains completely decoupled forces and torques that can be calculated independently.
This superiority is due to the six-dimensional force and torque reachability presented in \Cref{sec:dynamics}.

We first define tracking errors as 
\begin{equation}
    \begin{aligned}
        \boldsymbol{e}_{p} & = \boldsymbol{p} - \boldsymbol{p}_{d}, & \boldsymbol{e}_{R} & =\frac{1}{2}\left(\boldsymbol{R}_{d}^\top \boldsymbol{R} - \boldsymbol{R}^{\top} \boldsymbol{R}_{d} \right)^{\mathrm{\vee}}, \\
        \boldsymbol{e}_{v} & = \boldsymbol{v} - \boldsymbol{v}_{d}, & \boldsymbol{e}_{\omega} & = \boldsymbol{\omega} - \boldsymbol{R}^{\top} \boldsymbol{R}_{d} \boldsymbol{\omega}_{d};  \\
    \end{aligned}
\end{equation}
$\boldsymbol{p}_{d}$, $\boldsymbol{v}_{d}$, $\boldsymbol{R}_{d}$, $\boldsymbol{\omega}_{d}$ are desired position, velocity, rotation matrix, and angular velocity, and $\mathrm{\vee}: SO(3) \rightarrow \mathbb{R}^{3}$ is the vee map.

The generalized geometric controller calculates the desired force and torque as

\begin{equation}
    \begin{aligned}
        \boldsymbol{f}_{d} &= -\boldsymbol{R}^\top(-k_p \boldsymbol{e}_p - k_v \boldsymbol{e}_v -  m \boldsymbol{g}  + m \boldsymbol{a}_{d}), \\
        \boldsymbol{\tau}_{d} &= -k_R \boldsymbol{e}_R - k_\omega \boldsymbol{e}_\omega + \boldsymbol{\omega} \times \mathcal{J}\boldsymbol{\omega}\\
        &\quad\quad - \mathcal{J}\boldsymbol{\omega} \times \boldsymbol{R}^\top \boldsymbol{R}_{d} \boldsymbol{\omega}_{d} - \boldsymbol{R}^\top \boldsymbol{R}_{d} \dot{\boldsymbol{\omega}}_{d},\\
        \end{aligned}
    \label{eq:des_force_moment}
\end{equation}
where $\boldsymbol{a}_{d}$, $\dot{\boldsymbol{\omega}}_{d}$ are desired acceleration and angular acceleration, $k_p$, $k_v$, $k_{R}$, and $k_{\omega}$ are positive PD control gains.

Substitute the desired force and moment in \cref{eq:des_force_moment} to quadrotor dynamics in \cref{eq:translational_dynamics} and \cref{eq:rotational_dynamics} gives the corresponding closed loop system dynamics

\begin{equation}
    \begin{aligned}
        \dot{\boldsymbol{e}}_p &= \boldsymbol{e}_v, \\
        \dot{\boldsymbol{e}}_v &= \frac{1}{m} \left(-k_p \boldsymbol{e}_p - k_v \boldsymbol{e}_v \right), \\
        \dot{\boldsymbol{e}}_R &= \frac{1}{2} \left(\tr\left(\boldsymbol{R}^{\top} \boldsymbol{R}_{d} \right) I - \boldsymbol{R}^{\top} \boldsymbol{R}_{d} \right)\boldsymbol{e}_{\omega}, \\
        \mathcal{J}\dot{\boldsymbol{e}}_\omega &= -k_R \boldsymbol{e}_R - k_\omega \boldsymbol{e}_\omega. 
    \end{aligned}
    \label{eq:error_dynamics}
\end{equation}

\begin{proposition}[Exponential Stability \textit{Almost} Everywhere in $SO(3)$] \label{prop:exp_stable} 

Consider the controller defined in \cref{eq:des_force_moment}. Suppose that the initial conditions satisfy
\begin{equation}
    \begin{gathered}
        \Psi\left(\boldsymbol{R}(0), \boldsymbol{R}_{d}(0)\right) < 2, \\
        \left\|\boldsymbol{e}_{\omega}(0)\right\|^{2}<\frac{2 k_{R}}{\lambda_{\max }(\mathcal{J})} \left(2-\Psi\left(\boldsymbol{R}(0),\boldsymbol{R}_{d}(0)\right)\right),
    \end{gathered}
    \label{eq_ROA}
\end{equation}
where a real-valued error function $\Psi(\cdot,\cdot) : SO(3) \times SO(3) \rightarrow \mathbb{R}$ is defined as $\Psi\left(\boldsymbol{R}, \boldsymbol{R}_{d}\right) \triangleq \frac{1}{2} \tr\left(I - \boldsymbol{R}_{d}^{\top} \boldsymbol{R} \right)$.
Then, the origin of the closed-loop error dynamics in \cref{eq:error_dynamics} is exponentially stable.
\end{proposition}

We obtain exponential stability of the generalized geometric controller for almost every pair of $\left(\boldsymbol{R}(0), \boldsymbol{R}_{d}(0)\right).$ Specifically, \Cref{prop:exp_stable} holds for $\Psi\left(\boldsymbol{R}(0), \boldsymbol{R}_{d}(0)\right) < 2$, which indicates the initial attitude error between $\boldsymbol{R}(0)$ and $\boldsymbol{R}_{d}(0)$ should be less than $180\degree$. In other words, we prove that the closed-loop controller is exponentially stable in \textit{almost} entire $SO(3)$. This result improves the exponential stability of the geometric controller of a regular quadrotor, which requires initial attitude error between $\boldsymbol{R}(0)$ and $\boldsymbol{R}_{d}(0)$ should be less than $90\degree$~\cite{lee2010geometric}.

%% file: 4-inverse.tex
\section{Control Allocation}\label{sec:inverse}

We describe how we compute the low-level actuator commands $\alpha_i$, $\beta_i$ and $\Omega_i$ from desired force and torque commands computed from the generalized geometric controller. We present the inverse mapping of the dynamics given in \cref{eq:F_matrix} which is an under-determined system due to the over-actuated design of the quadrotor. Since this inverse mapping had infinite solutions, the proposed control allocation is provably energy optimal. Hence, our control allocation is energy optimal resulting in minimum thrust cancellations among the rotor arms. As described further, we provide an intuitive and computationally simple allocation scheme where the thrust vector for each arm consists of independent components for force, roll, pitch, and yaw commands. Note that, since each arm has vectored-thrust in any direction based on $\alpha_i$ and $\beta_i$. Using this, we can apply the maximum possible thrust uniformly in all directions, as seen in \Cref{fig:envelope}.

\begin{figure}[t]
    \centering
    \includegraphics[width=0.8\linewidth]{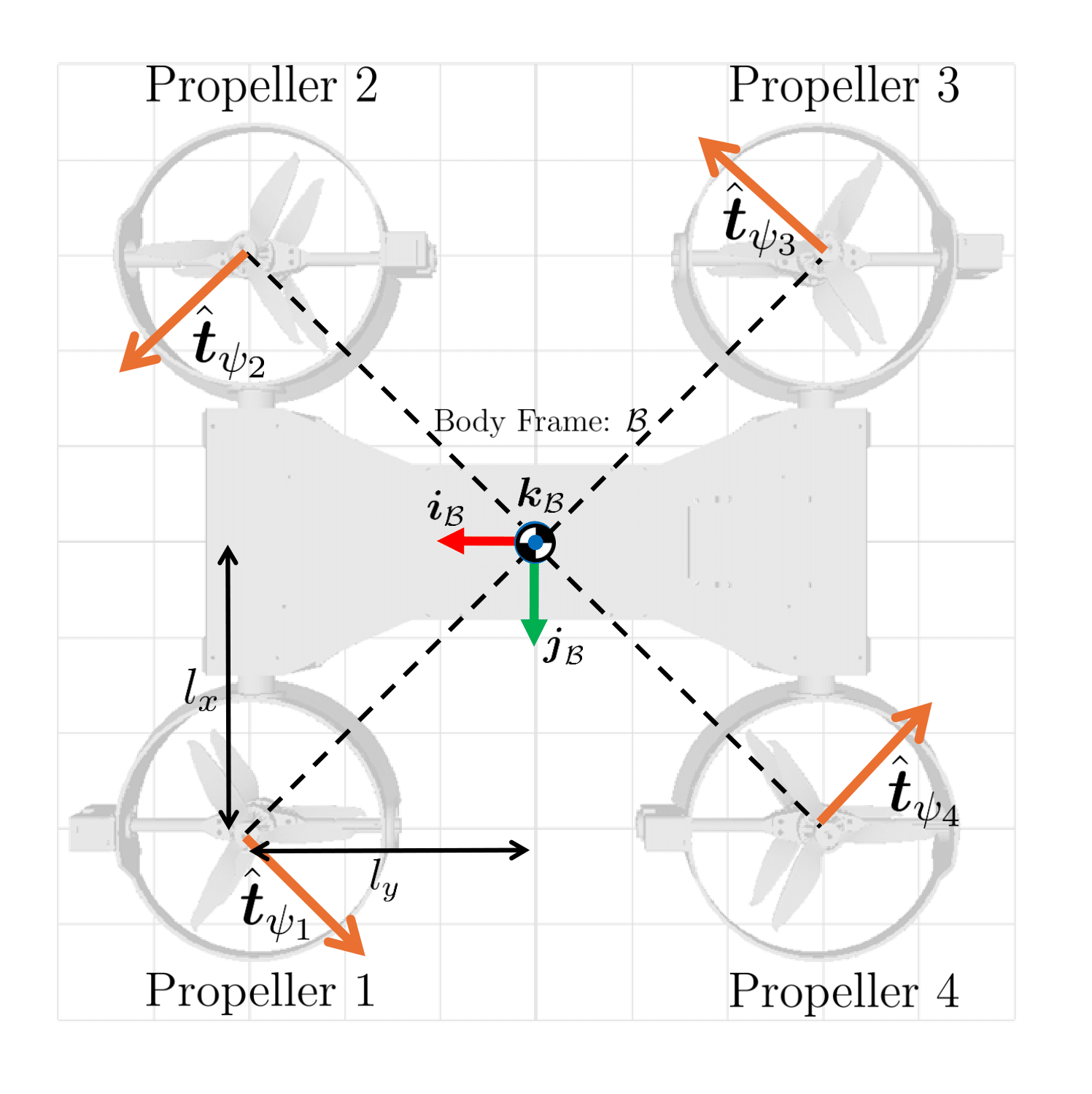}
    \caption{Unit vectors in $\boldsymbol{i}_{\mathcal{B}}-\boldsymbol{j}_{\mathcal{B}}$ for \textit{yaw} control.}
    \label{fig:inverse}
    \vspace{-4mm} 
\end{figure}

The desired force-torque vectors, $\boldsymbol{f}_d$ and $\boldsymbol{\tau}_d$, are computed using the geometric control law (\cref{eq:des_force_moment}). The low-level commands $\alpha_i$, $\beta_i$, and $\Omega_i$ are decoupled with the values for the other arm, hence the resulting thrust vector \(\boldsymbol{t}_i\) can be considered an (intermediate) virtual control input. Note that the magnitude $\|\boldsymbol{t}_{i}\|$ depends on the angular velocity of the propeller $\Omega_i$ and direction $\hat{\boldsymbol{t}}_{i}$ is determined by the values of servo motor angles $\alpha_i$ and $\beta_i$. Then we break down the desired force and torque into four components, \ie \textit{net force}, \textit{roll}, \textit{pitch} and \textit{yaw}. Each component is equally (in magnitude) distributed to all four arms/propellers such that it does not affect the other component. Mathematically, each thrust vector $\boldsymbol{t}_i$ is a vector sum of its contribution towards \textit{net force}, \textit{roll}, \textit{pitch} and \textit{yaw}, given by
\begin{equation}
\begin{aligned}
    \boldsymbol{t}_1 &= \frac{\boldsymbol{f}_d}{4} + \left(\frac{\tau_{x,d}}{4l_y}\right)\boldsymbol{k}_{\calB} + \left(\frac{-\tau_{y,d}}{4l_x}\right)\boldsymbol{k}_{\calB} + \left(\frac{\tau_{z,d}}{4r}\right)\hat{\boldsymbol{t}}_{\psi_1}, \\
    \boldsymbol{t}_2 &= \frac{\boldsymbol{f}_d}{4} + \left(\frac{-\tau_{x,d}}{4l_y}\right)\boldsymbol{k}_{\calB} + \left(\frac{-\tau_{y,d}}{4l_x}\right)\boldsymbol{k}_{\calB} + \left(\frac{\tau_{z,d}}{4r}\right)\hat{\boldsymbol{t}}_{\psi_2}, \\
    \boldsymbol{t}_3 &= \frac{\boldsymbol{f}_d}{4} + \left(\frac{-\tau_{x,d}}{4l_y}\right)\boldsymbol{k}_{\calB} + \left(\frac{\tau_{y,d}}{4l_x}\right)\boldsymbol{k}_{\calB} + \left(\frac{\tau_{z,d}}{4r}\right)\hat{\boldsymbol{t}}_{\psi_3}, \\
    \boldsymbol{t}_4 &= \frac{\boldsymbol{f}_d}{4} + \left(\frac{\tau_{x,d}}{4l_y}\right)\boldsymbol{k}_{\calB} + \left(\frac{\tau_{y,d}}{4l_x}\right)\boldsymbol{k}_{\calB} + \left(\frac{\tau_{z,d}}{4r}\right)\hat{\boldsymbol{t}}_{\psi_4},
\end{aligned} \label{eq:inverse}
\end{equation}
where the values $l_x$, $l_y$, and $r=\sqrt{l_x^2+l_y^2}$ are the half-length, half-breadth and distance from CoM of each propeller. This relation can be represented as a linear map $\calN:[\boldsymbol{f}_d, \boldsymbol{\tau}_d]^{\top}\in\mathbb{R}^6\rightarrow \boldsymbol{t}\in\mathbb{R}^{12}$. The vectors $\hat{\boldsymbol{t}}_{\psi_i}$ are the unit vectors along the yaw directions, in $\boldsymbol{i}_{\mathcal{B}}-\boldsymbol{j}_{\mathcal{B}}$ plane, as shown in \Cref{fig:inverse}. Mathematically, these are given by 
\begin{equation}
\begin{aligned}
    \hat{\boldsymbol{t}}_{\psi_1} =  \begin{bmatrix}
    \frac{-l_y}{r} & \frac{l_x}{r} & 0 \end{bmatrix}^\top, &\quad
    \hat{\boldsymbol{t}}_{\psi_2} =  \begin{bmatrix}
    \frac{l_y}{r} & \frac{l_x}{r} & 0 \end{bmatrix}^\top, \\
    \hat{\boldsymbol{t}}_{\psi_3} =  \begin{bmatrix}
    \frac{l_y}{r} & \frac{-l_x}{r} & 0 \end{bmatrix}^\top, &\quad
    \hat{\boldsymbol{t}}_{\psi_4} =  \begin{bmatrix}
    \frac{-l_y}{r} & \frac{-l_x}{r} & 0 \end{bmatrix}^\top.
\end{aligned}
\end{equation}

Finally, from the value of each thrust vector $\boldsymbol{t}_i$, we can now find the values of $\alpha_i$, $\beta_i$ and $\Omega_i$. Based on the rotation matrix $\boldsymbol{R}_{\mathcal{P}_i/\mathcal{B}}(\alpha_i,\beta_i) = \boldsymbol{R}_{\boldsymbol{j}_{\mathcal{P}_i}}(\alpha_i) \boldsymbol{R}_{\boldsymbol{i}_{\mathcal{P}_i}}(\beta_i)$, we have $\boldsymbol{t}_i$  as
\begin{align}
    \boldsymbol{t}_i = c_t \Omega_i^2 \begin{bmatrix} \sin\alpha_i \cos\beta_i \;\;  -\sin\beta_i \;\; \cos\alpha_i \cos\beta_i\end{bmatrix}^\top.
\end{align}

We can obtain multiple solutions ($\alpha_i,\beta_i$ values) for this equation. Due to angle wrapping for inverse trigonometric functions, by design, we define:
\begin{align}
    \Omega_i = \sqrt{||{t}_i||/c_t} ,\quad \;
     \beta_i = -\sin^{-1}\left(\hat{{t}}_{i,y}\right) ,\quad \;
\end{align}
\begin{align}
    \alpha_i =
\begin{cases} 
    \pi - \sin^{-1}\left(\frac{\hat{{t}}_{i,x}}{\cos\beta_i}\right) \in (\pi/2, \pi)  \\ \qquad \qquad \qquad \qquad \text{if } {t}_{i,z} < 0 \text{ and } {t}_{i,x} \geq 0, \\[5pt]
    -\pi - \sin^{-1}\left(\frac{\hat{{t}}_{i,x}}{\cos\beta_i}\right)\in (-\pi, -\pi/2) \\ \qquad \qquad \qquad \qquad \text{if } {t}_{i,z} < 0 \text{ and } {t}_{i,x} < 0, \\[5pt]
    \sin^{-1}\left(\frac{\hat{{t}}_{i,x}}{\cos\beta_i}\right)\in [-\pi/2, \pi/2] \\ \qquad \qquad \qquad \qquad \text{otherwise}.
\end{cases}
\end{align}
We can show that this control allocation design is optimal in the sense of sum-of-squares of the energy consumption (where energy is assumed to be proportional to $\Omega_i^2$). 
\begin{proposition}[Energy-efficient Control Allocation]
    The control allocation given by  \cref{eq:inverse} to find thrusts $\boldsymbol{t}_1,\boldsymbol{t}_2,\boldsymbol{t}_3,\boldsymbol{t}_4$,  for given $\boldsymbol{f}_d$ and $\boldsymbol{\tau}_d$, is the optimal solution with respect to the cost function $E(\boldsymbol{t}_i) = \frac{1}{2}\sum_{i=1}^4 \| \boldsymbol{t}_i\| ^2$.
\end{proposition}
This can be shown by solving the optimization for minimizing the above cost function subject to the constraints $\boldsymbol{f}=\sum_{i=1}^4 \boldsymbol{t}_i, \boldsymbol{\tau}=\sum_{i=1}^4 \boldsymbol{l}_i\times\boldsymbol{t}_i$. Lagrangian multiplier method can be used to show that the proposed inverse mapping (\cref{eq:inverse}) is the solution. Alternatively, we can show that the linear (inverse) map $\calN:[\boldsymbol{f}_d, \boldsymbol{\tau}_d]^{\top}\in\mathbb{R}^6\rightarrow \boldsymbol{t}\in\mathbb{R}^{12}$ is pseudo-inverse of forward dynamics mapping $\calM: \boldsymbol{t}\in \mathbb{R}^{12}\rightarrow [\boldsymbol{f},\boldsymbol{\tau}]^\top \in \mathbb{R}^6$, \textit{i.e.,} $N=M^{\dagger}=M^{T}\left(MM^{T}\right)^{-1}$. We know that this gives a minimum-norm squared solution for the under-determined system with infinite solutions.

%% file: 5-exp.tex
\section{Simulation Experiments}\label{sec:exp}

In this section, we present the experimental results demonstrating the proposed quadrotor design and the corresponding controller’s performance in inspection-inspired applications.

\subsection{Continuous Contact-based Inspection}

\begin{figure}[H]
    \centering
    \begin{minipage}{\linewidth}
        \centering 
        \includegraphics[width=0.7\linewidth]{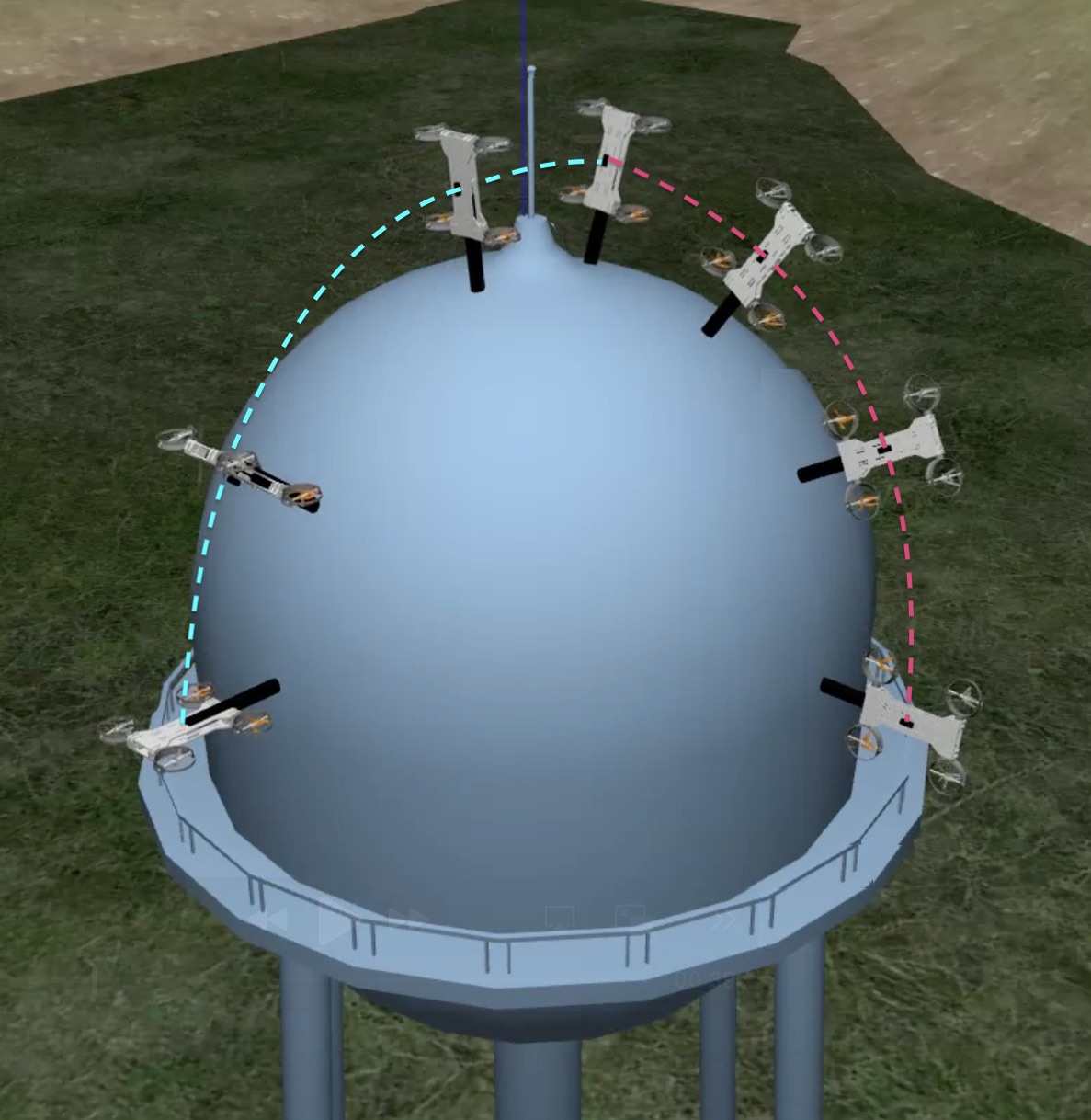}
    \end{minipage}
    \caption{Our quadrotor following a continuous-contact trajectory on a simulated water tower. During ascent (blue) and descent (red), it dynamically orients its fixed tool (black rod) to be normal to the structure.}
    \label{fig:watertower_qual}
\end{figure}

For the first experiment, we highlight a potential application of our drone design for contact-based inspection tasks. \Cref{fig:watertower_qual} demonstrates our vehicle’s ability to orient an attached tool in arbitrary directions with precision. In this experiment, we affix a black tool to the vehicle and trace a controlled trajectory along the surface of a water tower. As the vehicle ascends, it dynamically pitches to maintain proper tool alignment. Upon reaching the top, it transitions into a descent by yawing about its body z-axis while continuously keeping the tool in contact with the surface. This controlled interaction illustrates the system’s dexterity and ability to maintain precise tool orientation, expanding the range of tasks achievable with aerial robots beyond what is possible with conventional multirotors. 

\begin{figure}[H]
    \centering
    \begin{minipage}{\linewidth}
        \centering 
        \includegraphics[width=1\linewidth]{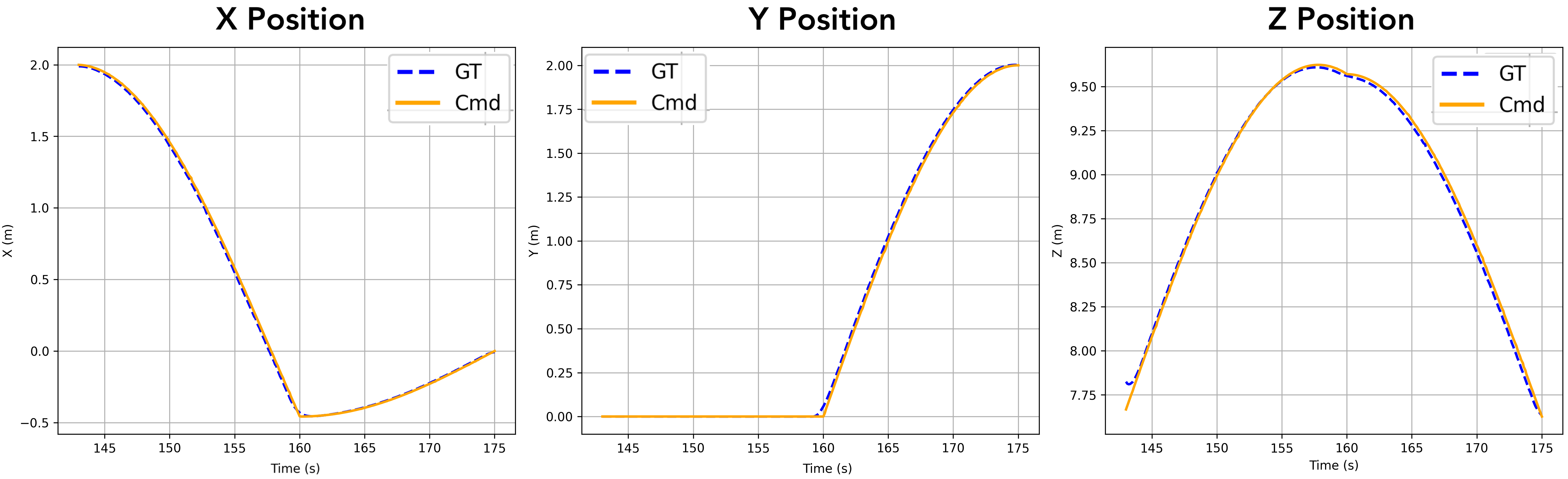}
    \end{minipage}

    \caption{Translational trajectories for commanded water tower trajectory. The vehicle position (blue dashed) closely follows the commanded position (orange solid).
    }
    \label{fig:trans_traj}
    \vspace{-4mm} 
\end{figure}

\begin{figure}[H]
    \centering
    \begin{minipage}{\linewidth}
        \centering 
        \includegraphics[width=1\linewidth]{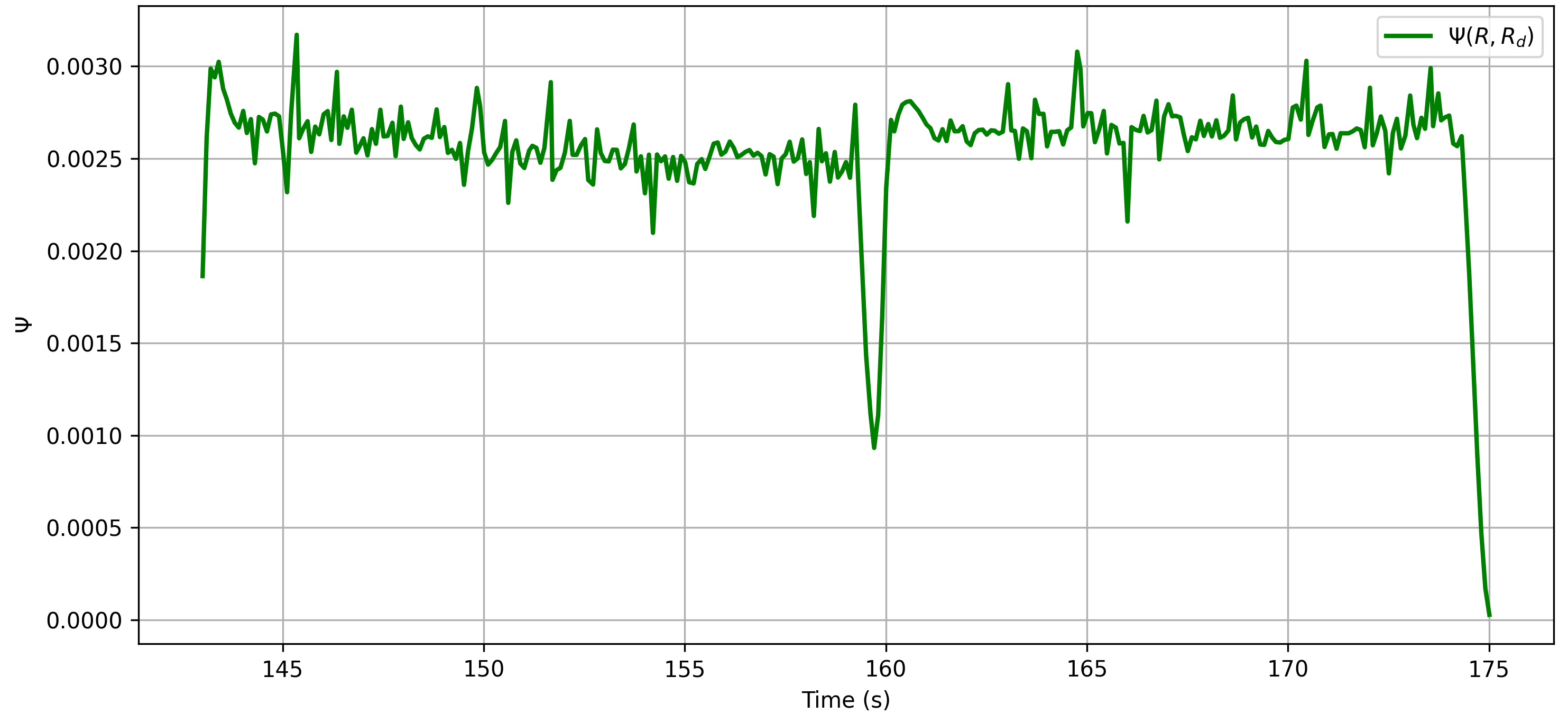}
    \end{minipage}

    \caption{Orientation error, $\Psi(R, R_d)$, for commanded water tower trajectory.}
    \label{fig:ori_err}
    \vspace{-4mm} 
\end{figure}

\Cref{fig:trans_traj} and \Cref{fig:ori_err} demonstrate our controller's effectiveness in tracking position and orientation during the maneuver. As shown in \Cref{fig:trans_traj}, the quadrotor's ground truth position precisely tracks the commanded position of the trajectory. Additionally, \Cref{fig:ori_err} indicates that throughout the entire maneuver, the orientation error never surpasses $0.0035$ radians. 

\subsection{Flying in Constrained Spaces}
\begin{figure}[H]
    \centering
    \begin{minipage}{\linewidth}
        \centering 
        \makebox[0pt][r]{\textbf{(a)}\quad} 
        \includegraphics[width=0.8\linewidth]{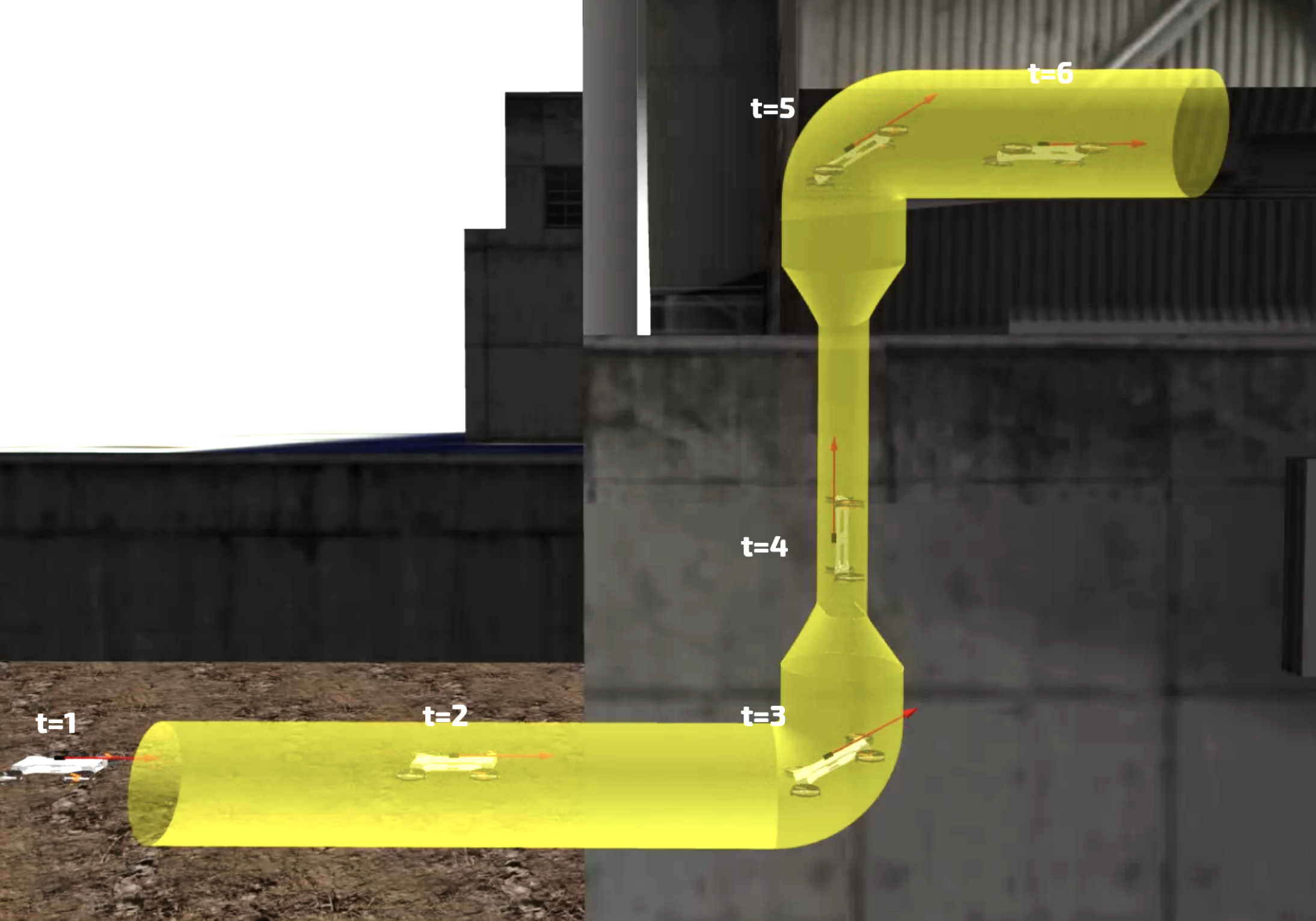}
    \end{minipage}

    \vspace{2pt} 

    \begin{minipage}{\linewidth}
        \centering 
        \makebox[0pt][r]{\textbf{(b)}\quad} 
        \includegraphics[width=0.8\linewidth]{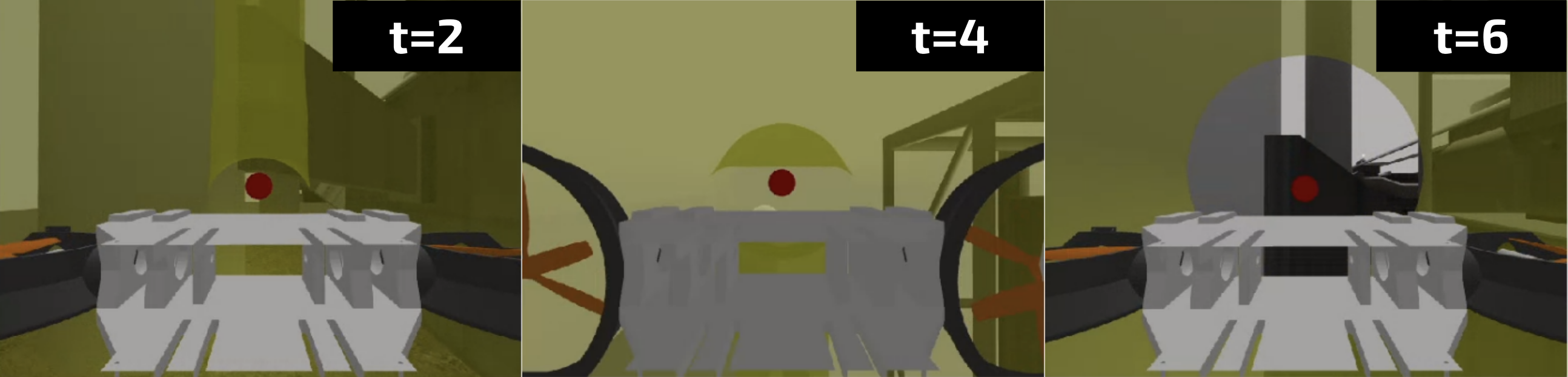}
    \end{minipage}

    \caption{\textbf{(a)} The morphable quadrotor tracking a trajectory inside a constrained conduit and \textbf{(b)} first-person-view inspection images from a body-mounted camera.}
    \label{fig:constrained}
\end{figure}

Next, we demonstrate our quadrotor’s ability to follow 6-DoF trajectories in constrained spaces.  \Cref{fig:constrained} highlights how the enhanced maneuverability of our quadrotor enables it to navigate successfully through a narrow pipe structure. The vertical section of the pipe imposes strict spatial constraints, limiting the feasible orientations that the quadrotor can achieve without collision. Our system's adaptability enables it to traverse these tight spaces.

\subsection{Continuous Visual Inspection}
\begin{figure}[H]
    \centering
    \begin{minipage}{\linewidth}
        \centering 
        \makebox[0pt][r]{\textbf{(a)}\quad} 
        \includegraphics[width=0.8\linewidth]{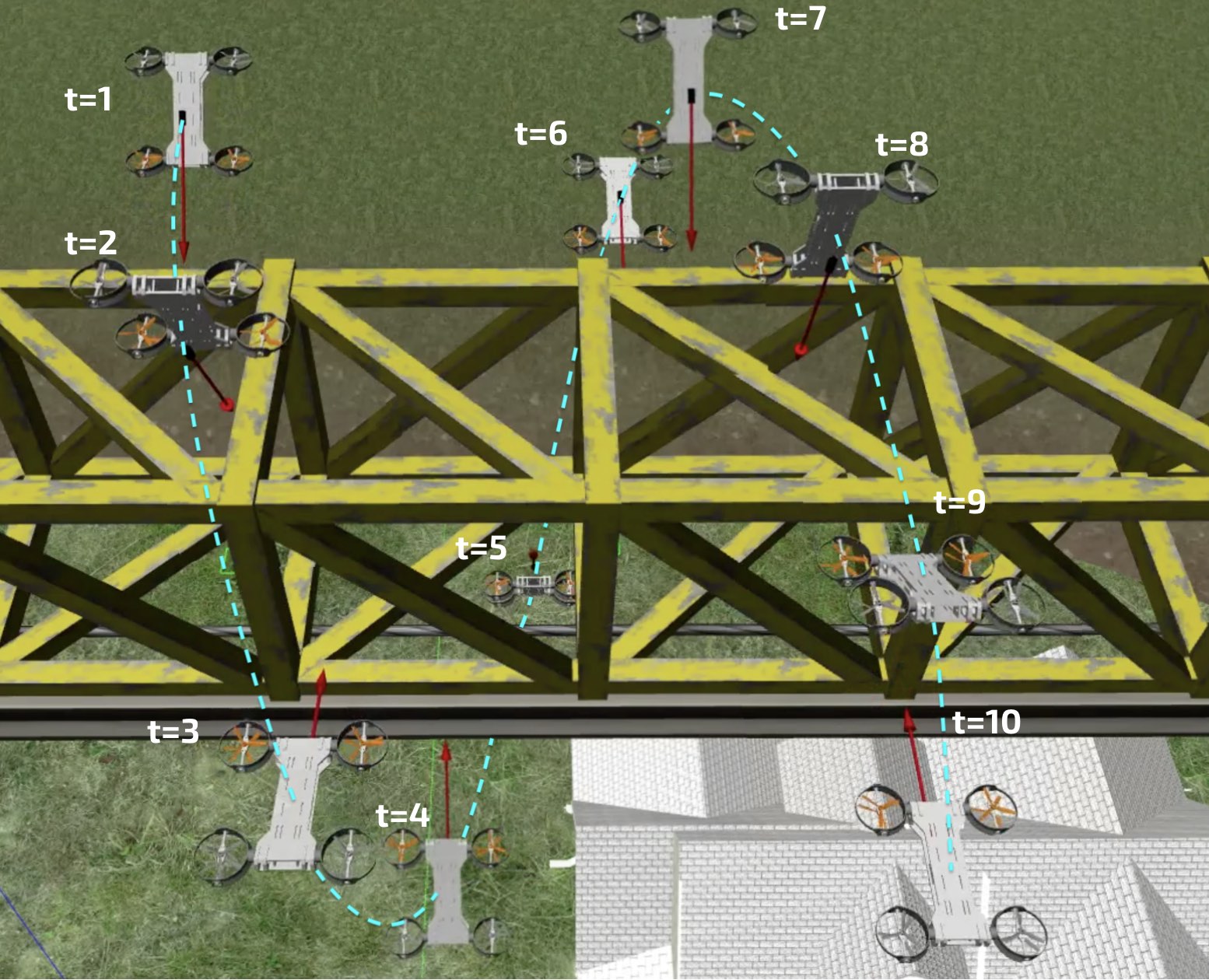}
    \end{minipage}

    \vspace{2pt} 

    \begin{minipage}{\linewidth}
        \centering 
        \makebox[0pt][r]{\textbf{(b)}\quad} 
        \includegraphics[width=0.8\linewidth]{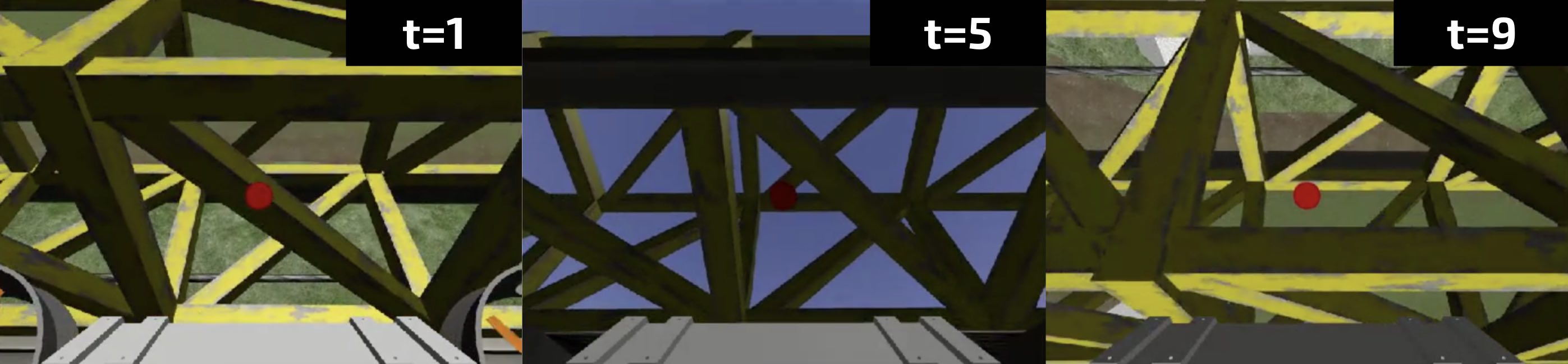}
    \end{minipage}

    \caption{\textbf{(a)} The morphable quadrotor tracking a corkscrew trajectory (dotted blue line) around an object for 10 distinct timesteps and \textbf{(b)} the corresponding first-person-view inspection images from a body-mounted camera. The red vector represents the optical axis of the camera.}
    \label{fig:corkscrew}
    \vspace{-4mm} 
\end{figure}

Finally, to showcase the system's 6-DoF trajectory-following capabilities, we design an experiment in which the quadcopter follows a corkscrew trajectory while continuously keeping a fixed camera aimed at a central object. Unlike current multirotors, which are limited in their ability to maintain precise viewpoint control during complex maneuvers, our system leverages its extended maneuverability to smoothly follow the trajectory while ensuring the target remains consistently framed.  \Cref{fig:corkscrew} presents both a third-person perspective of the trajectory and a corresponding first-person view from the body-fixed camera.

%% file: 6-conclusion.tex
\section{Conclusion}\label{sec:con}
We presented a novel variable-tilt quadrotor and accompanying control pipeline that achieves high levels of maneuverability and hand-like dexterity. By independently actuating each rotor's orientation, our vehicle enables optimally energy-efficient 6-DoF control of position and orientation, enabling the vehicle to stably achieve almost any configuration in $SO(3)$. We demonstrate applications of this quadrotor in simulations: (i) maintaining continuous contact with surfaces during flight, (ii) flying in constrained spaces, and (iii) continuous vision-based inspection. 

\begin{figure}[h]
    \centering
    \includegraphics[width=0.40\textwidth]{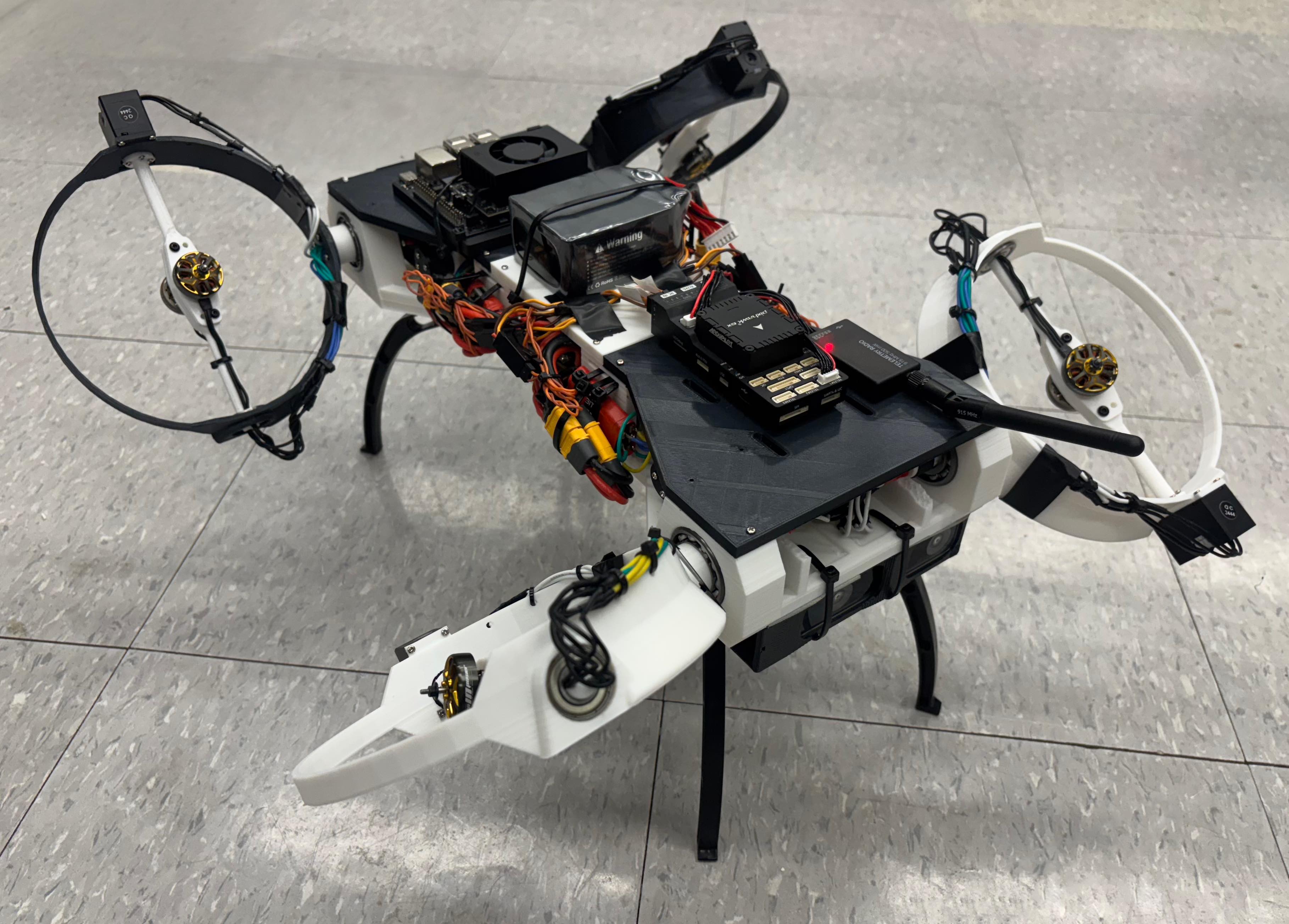}
    \caption{Prototype of the morphable quadrotor.}
    \label{fig:real_world}
\end{figure}

In future work, we plan to apply these algorithms in real-world experiments. Progress has already been made in this direction, including: (i) extending the algorithms herein to the PX4~\cite{px4} software-in-the-loop flight stack, (ii) fabricating the drone (\Cref{fig:real_world}), and (iii) conducting preliminary flight tests. For more information, please visit our project website.\footnote{\url{https://iral-morphable.github.io/}}

